\documentclass[12pt]{article}

\usepackage{amssymb}
\usepackage{makeidx}  
\usepackage{amsmath}
\usepackage{hyperref}
\usepackage{epsfig}
\usepackage{subfigure}
\usepackage{graphicx}
\usepackage[round]{natbib}
\usepackage{multirow}
\usepackage{algpseudocode}
\usepackage[autostyle]{csquotes}
\usepackage{color}
\graphicspath{{Paper_images/}}

\begin{document}

\title{Recurrent Convolutional Networks for Pulmonary Nodule Detection in CT Imaging}


\author{Petros-Pavlos Ypsilantis \qquad Giovanni Montana\\Department of Biomedical Engineering
\\King's College London}

\date{}


\maketitle

\begin{abstract}

Computed tomography (CT) generates a stack of cross-sectional images covering a region of the body. The visual assessment of these images for the identification of potential abnormalities is a challenging and time consuming task due to the large amount of information that needs to be processed. In this article we propose a deep artificial neural network architecture, ReCTnet, for the fully-automated detection of pulmonary nodules in CT scans. The architecture learns to distinguish nodules and normal structures at the pixel level and generates three-dimensional probability maps highlighting areas that are likely to harbour the objects of interest. Convolutional and recurrent layers are combined to learn expressive image representations exploiting the spatial dependencies across axial slices. We demonstrate that leveraging intra-slice dependencies substantially increases the sensitivity to detect pulmonary nodules without inflating the false positive rate. On the publicly available LIDC/IDRI dataset consisting of 1,018 annotated CT scans, ReCTnet reaches a detection sensitivity of 90.5$\%$ with an average of 4.5 false positives per scan. Comparisons with a competing multi-channel convolutional neural network for multi-slice segmentation and other published methodologies using the same dataset provide evidence that ReCTnet offers significant performance gains.  

\end{abstract}

\section{Introduction}

Lung cancer is the second most common cancer in both men and women in Europe \citep{Sant09} and in the United States \citep{Acs15}. In the UK, it is the second most common malignancy, with 44,500 new cases of the disease in 2012, and the most common cause of cancer-related death with over 35,000 deaths in 2012 alone \citep{cr15}. Survival with the disease is the second lowest across all malignancies with just over $30\%$ surviving one year post-diagnosis, and only $5\%$ surviving ten years post-diagnosis \citep{cr15}. This is in great part due to the late diagnosis and commencement of appropriate management plans. The later the disease is diagnosed and managed, the poorer the outcome is for the patient, hence the requirement for as early and as accurate an intervention as possible. Recent advances in computed tomography (CT) imaging have resulted in early diagnosis of the disease \citep{Li07}. For instance, it has been reported that the detection rate of lung cancer using CT is $2.6$ to $10$ times higher compared to chest radiography \citep{Sobue02, Sone98, Henschke99, Sone00}.


CT scanners produce up to $600$ cross-sectional 2D images that must be individually evaluated in a short time. Despite the diagnostic benefits provided by CT imaging, the increased workload that is required to read hundreds of slices per exam can lead to a larger number of mistakes being made. Existing studies have shown that the rate of erroneous CT interpretation increases from $7\%$ to $15\%$ when a radiologist performs more than $20$ CT examinations per day \citep{Bechtold97}. In order to address these issues, in the last few years there has been a burst of research activity around the development of computer-aided diagnosis (CAD) systems for pulmonary nodule detection using CT imaging \citep{Lee01, Shah05, Li07, Murphy09}. The adoption of CAD systems that facilitate CT reading has been proved to significantly improve the sensitivity to detect nodules with diameter larger than $3$mm from $44\%$ to $57\%$ in \cite{Bogoni12} and from $56\%$ to $67\%$  in \cite{Sahiner09}.

A large number of recently published image analysis approaches for pulmonary nodule detection using CT imaging employ machine learning at their core, and share a common two-stages process \citep{Golosio08, Messay10, Tan11, Torres15, Teramoto12, Opfer07, Liu10}. First, a region generating mechanism is deployed to produce a large number of candidate regions having high likelihood of containing the pulmonary nodules. These candidate regions are typically identified using intensity-based and/or morphology-based imaging features, and often rely upon prior assumptions about the expected morphology of the nodules. In a second stage, a feature vector is extracted from the candidate regions and used to train a statistical classifier in an attempt to reduce the number of false positives resulting from the first stage. The main limitation of such a two-stage approach is that inaccurate assumptions about the underlying morphological and other appearance characteristics of the anatomical structures can result in low sensitivity rate in the first stage. For example, it is commonly assumed that nodules have spherical appearances \citep{Li04, Lee01} whereas a wider variety of abnormal anatomical structures with broader morphological variety can in fact be observed \citep{Li02, White96}. Clearly, training a classifier on imaging features that are not sufficiently discriminative can result in low detection performance.  

The aim of this article is to explore the feasibility of deep artificial neural networks for pulmonary object detection. In recent years, deep learning has emerged as a powerful approach to learning imaging representations directly from large volumes of data thus dispensing from the need to hand-engineer predictive features \citep{Bengio14, Hinton07}. Convolutional neural networks (CNNs) have been shown to be able to learn hierarchically organised low to high-level features directly from raw images \citep{Fukushima80, LeCun98}, and yield state-of-the-art performance in 2D image classification \citep{Szegedy16}, object detection \citep{Ren15, He15} and semantic segmentation \citep{Long15, Yu16} tasks. In object detection the representational power of convolutional feature maps is exploited to generate region proposals and guide the search for object of interests, thereby avoiding exhaustive sliding window searches across images \citep{Girshick13}. For semantic segmentation tasks, dilated convolutions have been utilized to systematically expand the receptive field of convolutional layers and aggregate contextual information without losing resolution \citep{Simonyan15}. These methods have shown promising results on natural images where the objects to be detected and segmented are typically well-defined and sufficiently large compared to the entire image. However, the majority of nodule volumes typically cover less than $0.01\%$ of the lung area, and there is often a striking similarity between nodular and normal structures as seen on 2D slices, e.g. when the nodules are attached to normal pulmonary structures and pleura. 

\begin{figure}[t]
\begin{center}
\includegraphics[width=5.0in]{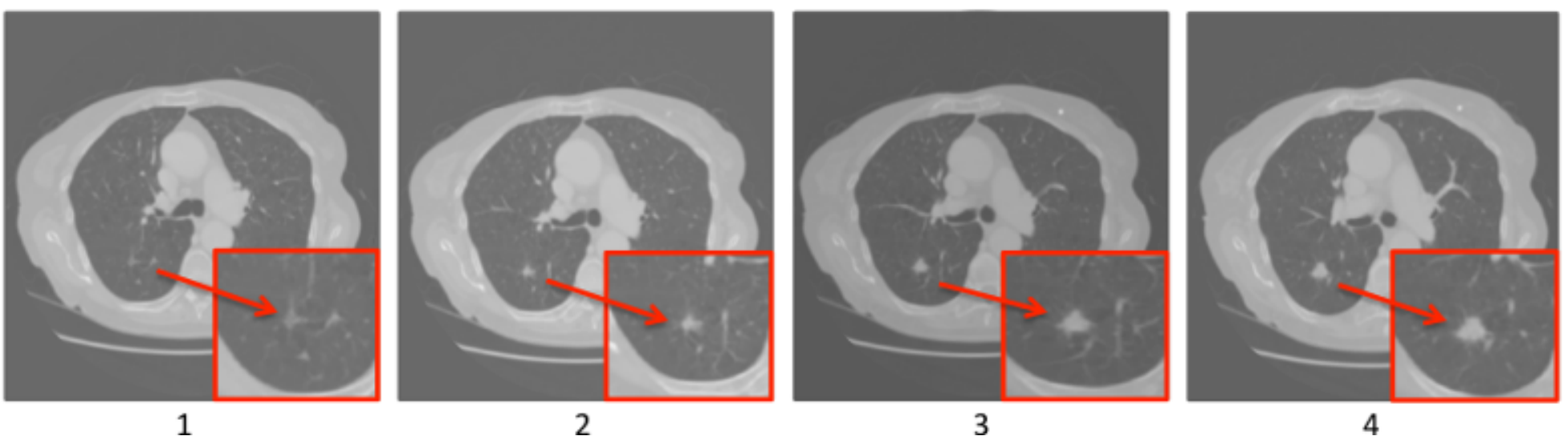}
\end{center}
\caption{Internal anatomical structure of a pulmonary nodule as shown in four transverse consecutive sections of a CT scan. The morphological characteristics of the nodule change gradually across slices.}
\label{fig:nodule_sequence}
\end{figure}

In this work we set out to mimic the reading process carried out by radiologists, who would typically explore all the slices in the stack and draw inferences about the likely presence of a nodule by picking up changes in radiological appearance between adjacent slices. The need for integrating inter-slice information for nodule detection and characterisation has often been emphasised in the literature \citep{Golosio08, Tan11, Messay10}. As an illustrative example of this, Figure~\ref{fig:nodule_sequence} shows four adjacent cross-sectional slices extracted from a thoracic CT scan. All images contain a pulmonary nodule whose morphological characteristics differentiate it from the surrounding normal tissue. Our hypothesis is that learning imaging features that capture such inter-slice spatial correlations can eventually introduce performance gains in nodule detection. In order to address this problem, we propose an artificial neural network that combines the representational power of CNNs with the capability to learn dependencies in sequential data that is typical of recurrent neural networks (RNNs) \citep{Hochreiter97}. RNNs have achieved tremendous success in solving challenging sequential problems like machine translation and speech recognition \citep{Bahdanau14, Graves14, Sutskever14}. More recently, multi-dimensional RNNs have shown state-of-the-art performance in scene lebeling in 2D images \citep{Byeon14}, volumetric image segmentation \citep{Stollenga15, Poudel16} and modeling the distribution of natural images \citep{Oord16}.

The hybrid CNN-RNN architecture we propose, called ReCTnet, operates at the pixel level, and combines convolutional layers, which are able to learn discriminative features from raw scans, with recurrent layers, which facilitates learning anatomical dependencies across adjacent images in the entire stack of cross sectional images. As such, ReCTnet learns both within- and across-images features thus taking into full account the anatomical context of each pixel. Training is carried out end-to-end so that both the convolutional and recurrent components are learned jointly. Three-dimensional probability maps are generated as output indicating the likelihood of a nodule being present in certain areas. To the best of our knowledge such as a hybrid system has never been utilized for object detection in 3D imaging. 

\section{Material and methods} \label{methods}
 
\subsection{LIDC/IDRI dataset} \label{data}
 
For this study we have used CT imaging data generated by the Lung Image Database Consortium (LIDC) and Image Database Resource Initiative (IDRI) \citep{Armato11, Clark13}. The LIDC/IDRI database is the result of a joint effort between seven academic institutions and eight medical imaging companies supporting the development, training, and evaluation of CAD methods for lung cancer detection and diagnosis. As of February 2015, the database contained $1018$ CT scans acquired using single and multi-slice scanners with slice thickness ranging from $0.6$ mm to $5$ mm. The reconstruction interval ranges from $0.45$ mm to $5.0 $ mm and the in-plane pixel size ranges from $0.461$ mm to $0.977$ mm. 

\begin{figure}[t]
\begin{center}
\includegraphics[width=5.2in]{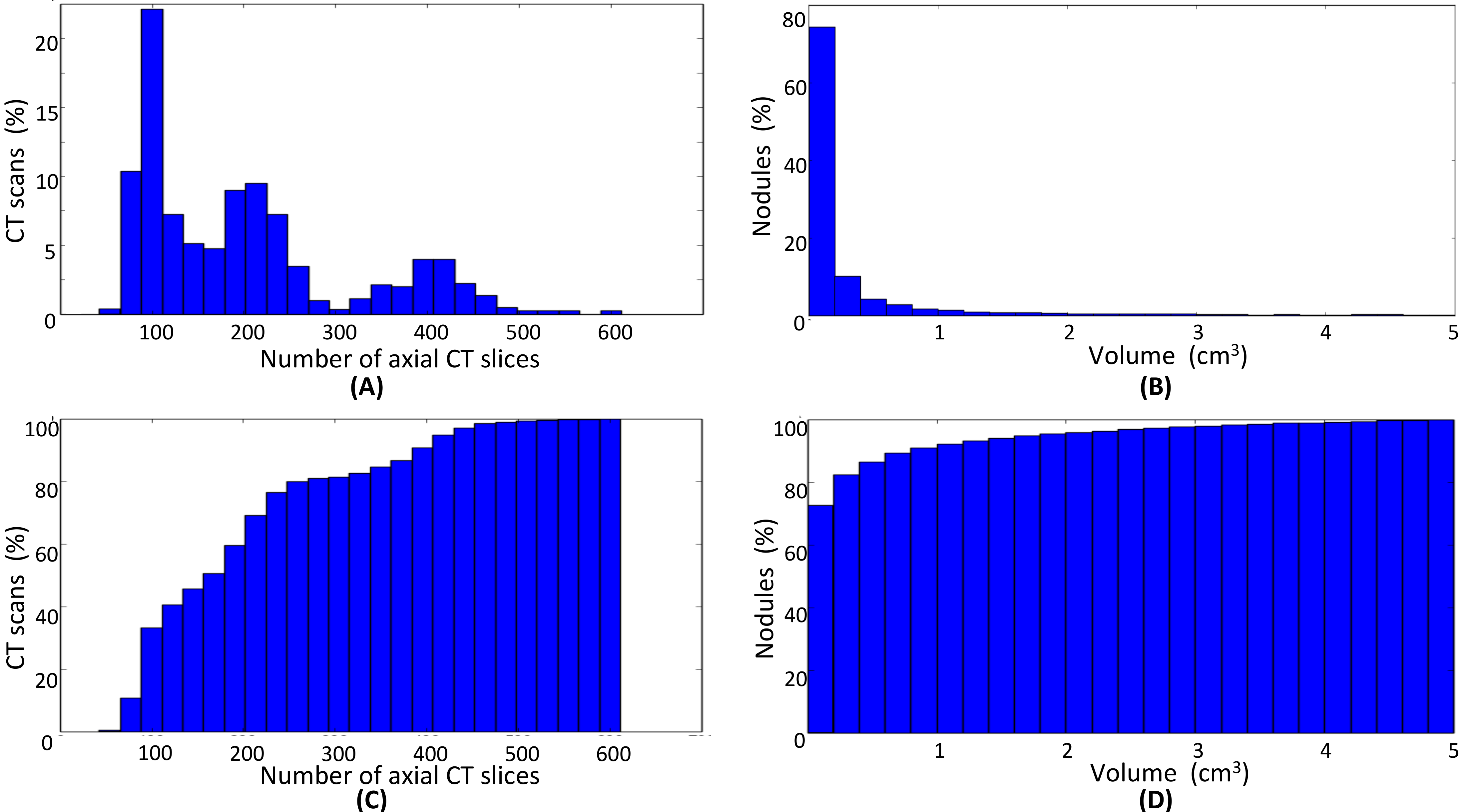}
\end{center}
\caption{Variability of nodule size and number of axial slices per CT scan in the LIDC/IDRI database: (A) frequency distribution of number of axial slices per CT scan, (B) frequency distribution of nodule volume size in cm$^{3}$, (C) cumulative frequency distribution of number of axial slices per scan, (D) cumulative frequency distribution of nodule volume size in cm$^{3}$.}
\label{fig:nodule_sizes}
\end{figure}

Each scan is provided along with annotations independently generated by four experienced radiologists affiliated with different institutions. Complete outlines for all nodules between $3$mm and $30$mm in diameter were obtained alongside with their radiological characteristics. The variability of nodule size is likely to depend on several factors, such as the characteristics of the nodule's attenuation and its location, the scan acquisition parameters, the quality of segmentation and the inter-scan and inter-reader variability \citep{Petkovska07}. Figure~\ref{fig:nodule_sizes} illustrates the variability of the number of axial slices per CT scan and the nodule size in the LIDC/IDRI database. An initial blinded review, in which each radiologist read the scan independently, was followed by a unblinded session in which each radiologist re-examined all the cases upon disclosure of the reports generated by the other readers. A varying level of disagreement among radiologists was observed for several findings after the unblinded reading phase, and no forced consensus was imposed in the final review. For the purpose of this study, we designate a voxel to belong to a nodule if the voxel falls within the contours drawn by at least one of the four radiologists.

For nodules with a diameter $\ge 3$ mm, the radiologists also provided ratings for subtlety and likelihood of malignancy, both on a scale from $1$ to $5$.  The subtlety score refers to the difficulty of detecting nodules by visual inspection only, with lower scores indicating greater difficulty. The malignancy score was obtained by taking into account the patient's age and smoking history, and lower scores indicate a smaller likelihood of malignancy. In our experiments, first we averaged both the subtlety and malignancy ratings across all four radiologists, and then grouped the scores into three classes: a nodule was labelled as having low, medium and high malignancy depending on whether the average score was below $2.5$, between $2.5$ and $3.5$ and above $3.5$, respectively. Analogously, a nodule's subtlety was classified as difficult, medium and easy depending on the average subtlety score.

\subsection{Methods} \label{methods}

This section details the proposed neural network architecture. In Section \ref{lung_detection} we describe a preprocessing algorithm for the segmentation of the lung area. Although not strictly necessary, the step was performed in our study to exclude portions of the images that are not expected to present pulmonary nodules. This level of preprocessing can help reduce the number of false positives and keep the computational cost during training more manageable. In Section \ref{sampling1} we detail the procedure followed to prepare training, validation and testing datasets, which are then used in all the experiments presented in Section \ref{results}. The architectural components of our deep neural network, ReCTnet, are presented in Section \ref{model} along with an alternative architecture that is solely using convolutional layers for comparative purposes. The output of these architectures consists of three-dimensional probability maps indicating the likelihood that each voxel belongs to nodular region. Finally, in Section \ref{validation}, we propose a post-processing procedure for clustering high-probability voxels that are spatially close to each other into homogeneous regions forming candidate nodules. 
 
\subsubsection{Lung area segmentation} \label{lung_detection}

All CT scans are initially pre-processed in order to identify and extract the lung areas, which will be used as input for training the neural network described in Section \ref{model}. For each CT scan in the database, the axial slices containing the lungs are represented by a tensor, $\mathcal{I}_{s} := [{\bf I}_{1}, \hdots, {\bf I}_{N_{s}}]$, where each ${\bf I}_{j} \in \mathcal{R}^{512 \times 512}$ is a cross-sectional slice in which the internal anatomy of the lungs and relevant body parts are depicted, and $s \in \{1, \ldots, 1018\}$. The number of slices per scan, $N_s$, depends on the scan index and varies from a minimum of $43$ to a maximum of $611$. 

\begin{figure}[t]
\begin{center}
\includegraphics[width=5.0in]{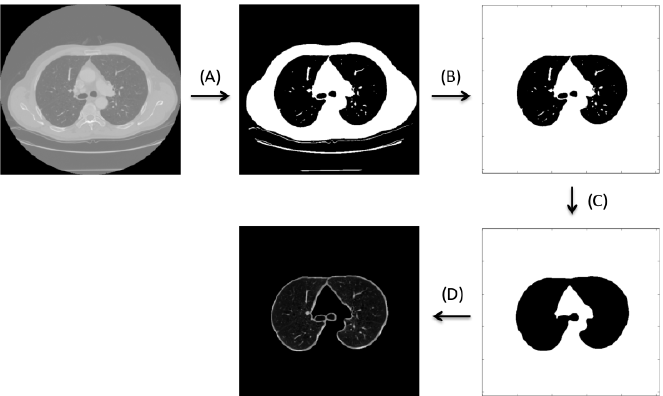}
\end{center}
\caption{Lung wall identification on a single CT slice: (A) The CT image is binarised using a threshold value of $-480$ HU. (B) A flood filling strategy is used to remove the non-lung regions after thresholding. (C) A morphological dilation operation is applied on the binary mask to form a new binary mask which includes the lung area and the pleural layer. (D) The new binary mask is applied to the original CT slice.}
\label{fig:lung_segmentation}
\end{figure}

Since the intensity in the original CT slices is proportional to tissue density, the lung areas can be segmented out through a thresholding-based region filling strategy \citep{Ko03}. The various steps of this strategy are illustrated in Figure~\ref{fig:lung_segmentation}. The pixel intensity in lung tissue varies approximately from $-400$ to $-600$ Hounsfield Units (HU), while the intensity for chest wall, body, and bone is found to be above $-100$ HU \citep{Wu94}. We have found that using a threshold of $-480$ HU for all voxels in each slice results in a robust segmentation of the lungs. A $2$D flood fill operation is then performed on each slice to remove non-lung regions left after thresholding and fill the resulting binary mask. Finally, a morphological dilation operation \citep{Soille03} is applied to form a new binary mask that includes the lung area and the pleural layer where the juxtapleural nodules are to be found. 

\subsubsection{Classification of voxels through local patches}\label{sampling1}

\begin{figure}[t]
\begin{center}
\includegraphics[width=5.2in]{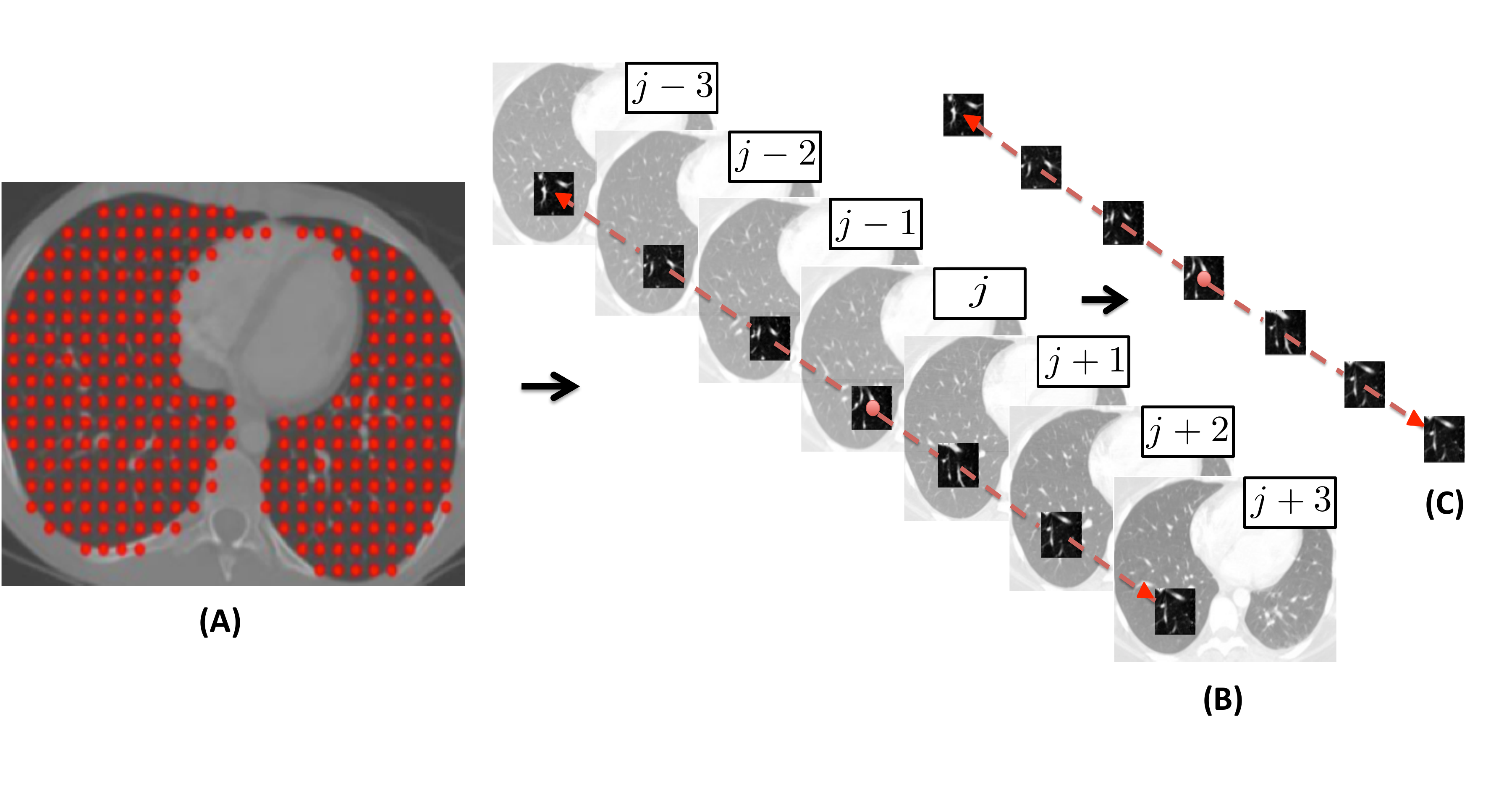}
\end{center}
\caption{(A) 2D uniform sampling grid of the $j$-th axial slice. The red points correspond to sampled voxels, all in the lung area; (B) For a sampled voxel in the $j$-th axial slice, we extract image patches from the $3$ slices above and $3$ slices below the $j$-th slice, including the central one, all sharing the same $(x,y)$ coordinates; (C) The resulting image stack containing the contextual information for the target voxel.}
\label{fig:uni_sampling}
\end{figure}

Our objective is to build a classifier that discriminates between voxels belonging to normal vascular tissue and those belonging to nodular areas. For this purpose, we need to set up a training dataset consisting of examples representative of both anatomical areas. Making use of all the available voxels was not computationally feasible and the two classes are highly unbalanced: overall there are approximately $12$ billion voxels representative of normal tissue and only $2.2$ million voxels from nodular areas. We have chosen to follow a sampling strategy whereby a smaller and more balanced number of voxels is pre-selected from the segmented lung areas. As illustrated in Figure~\ref{fig:uni_sampling} (A), on each slice we apply a uniform sampling grid with sampling distance along both directions taken to be $25$ times the in-plane pixel length of the corresponding CT scan. Using this grid, on average we sample $108$ voxels per slice. Each sampled voxel $c$ in slice ${\bf I}_{j}$ is associated with a corresponding binary class label, $y_{c,j}$, which is set to $1$ for voxels belonging to a nodule, otherwise equals $0$. Voxels that fall within a nodule \enquote{ground truth} are sampled randomly at a higher spatial resolution in order to explore and learn more thoroughly their anatomical characteristics.

In order to capture the local anatomical context around a given voxel $c$, we build a stack of small rectangular patches centered around $c$ and spanning a fixed number of axial slices just below and above the slice containing $c$. We call ${\bf X}_{c,j} \in \mathcal{R}^{M \times M}$ the patch extracted from the slice indexed by $j$ and containing the target voxel $c$. The stack is then represented by a tensor $\mathcal{X}_{c,j} := [{\bf X}_{c,j-k}, \hdots, {\bf X}_{c,j}, \hdots, {\bf X}_{c,j+k}]$, where all patches share the same $(x,y)$ coordinates. Each stack is taken to represent the context surrounding a voxel, and the process is illustrated in Figure~\ref{fig:uni_sampling} (B) and \ref{fig:uni_sampling} (C). In all the experiments presented here we have taken $k=3$ yielding a total of $7$ slices per stack; this has been found to be optimal for this application. Furthermore, two strategies have been implemented to build an augmented set of examples with a view on preventing overfiting and increasing the network's ability to generalise on unseen examples. First, we have adopted a multi-scale approach (e.g. \cite{Krizhevsky12}, \cite{Zeiler14}) to represent the anatomical context around a voxel at two different scales. This has been achieved by taking patches of two different sizes, i.e. $M=50$ and $M=80$. The larger patches are subsequently scaled down to match the smaller size. Moreover, as in previous work  (e.g. \cite{Krizhevsky12,Ciresan13}), we have augmented the training dataset by creating new examples from existing ones. This was achieved by simultaneously flipping and rotating all patches within a stack. The resulting training dataset consists of $17$ and $1.7$ million tensors, all of size $7 \times 50 \times 50$, representative of non-nodular and nodule areas, respectively.

\subsubsection{A recurrent convolutional neural network architecture}\label{model}

\begin{figure}[t]
\begin{center}
\includegraphics[width=5.5in]{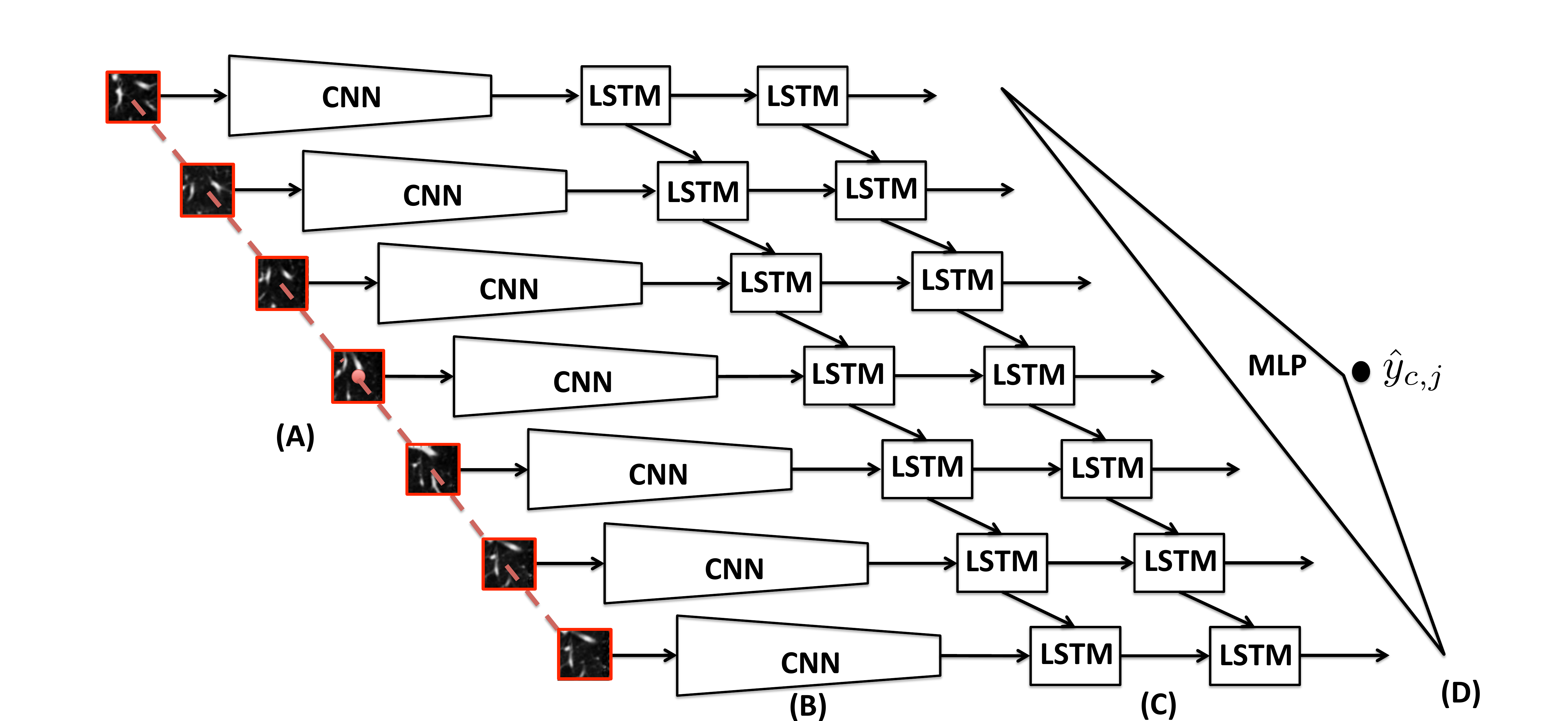}
\end{center}
\caption{An overview of the ReCTnet architecture for object detection in multi-slice medical images; (A) Adjacent CT image patches of the tensor $\mathcal{X}_{c,j}$. The red point in the middle slice of the sequence represents the voxel of interest. (B) CNNs for feature extraction; (C) Stack of $2$ LSTM layers composed of LSTM neurons for learning the anatomical dependencies across the adjacent image patches; (D) The output of the last LSTM layer is concatenated and passed as input to a fully connected multi-layer perceptron (MLP) that predicts the voxel label.}
\label{fig:general_model}
\end{figure}

In this section we introduce the ReCTnet architecture as illustrated in Figure~\ref{fig:general_model}. Initially, a CNN component learns a representation for each image patch, individually taken, within a stack. This component consists of a sequence of convolutional and max-pooling layers \citep{Nagi11}. Each neuron in a convolutional layer receives information from only a subset of the input image, its receptive field. As a result, each neuron learns to detect a particular pattern from a local region of the input image. This local connectivity captures the local substructure and preserves the topology of the input image \citep{Lecun89}. In addition to local connectivity, the convolutional layer also imposes groups of neurons, called feature maps, to share identical weight values. Weight sharing across groups of neurons reduces the number of free parameters thus increasing the generalization ability of the network \citep{Lecun89}. The output values of the neurons in the $m$-th feature map of the $l$-th convolutional layer is given by
\begin{equation}
{\bf v}_{m}^{(l)} = f(\sum_{s} {\bf K}_{s,m}^{(l)} \ast {\bf v}_{s}^{(l-1)} + b^{(l)}_{m} ),
\end{equation}
where $f$ is a non-linear activation function, $\ast$ is the convolutional operation and ${\bf K}_{s,m}^{(l)}$ is the weight matrix for the $m$-th feature map of the $l$-th layer and $s$-th feature map of the $l-1$-th layer.  ${\bf v}_{s}^{(l-1)}$ is the $s$-th feature map of the previous $l-1$-th layer and $b^{(l)}_{m}$ is the scalar bias of the $m$-th feature map of the $l$-th layer. 

Each convolutional layer consists of several feature maps so that a rich variety of features can be extracted at each location of the input image. The role of the max-pooling layer in each CNN component is to reduce the dimensionality of the feature maps \citep{Nagi11}. This is achieved by retaining only the maximum value within each non-overlapping sub-region of size ($a \times a$) for each feature map. Max-pooling layers have been shown to improve the generalization performance of CNN by selecting superior invariant features \citep{Nagi11}. Each CNN component in ReCTnet is obtained by stacking convolutional and max-pooling layers thus creating a multi-layer non-linear mapping $F({\bf X}_{c,j}| {\bf K})$, parameterized by ${\bf K}$ with $F: {\bf X}_{c,j} \to {\bf v}_{c,j}$ mapping each element of the tensor ${\bf X}_{c,j}$ into a fixed length representation ${\bf v}_{c,j}$. Here ${\bf K}$ represents the weights of all the convolutional operations across the entire CNN and ${\bf v}_{c,j} \in \mathcal{R}^{Q}$ is a vector of size $Q$ that encloses spatial characteristics associated with the anatomical structure depicted in the sequence of adjacent image patches. 

\begin{figure}[t]
\begin{center}
\includegraphics[width=5.0in]{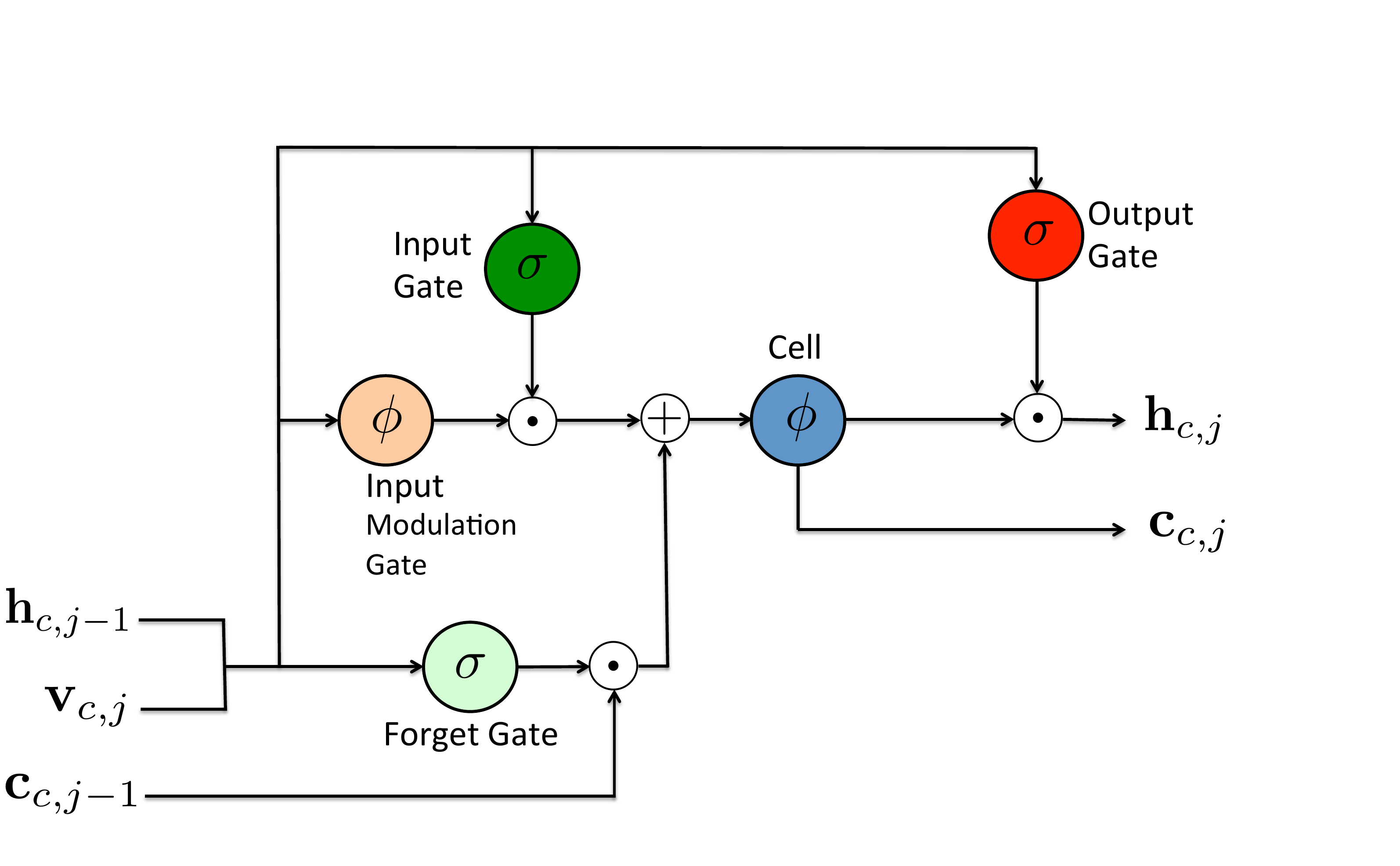}
\end{center}
\caption{Long short term memory neuron consisting of one memory cell. The gating mechanism controls the information flow across the entire sequence by independently reading, writing and erasing contents from the memory cell.}
\label{fig:lstm}
\end{figure}

The output of each CNN module provides a high-level representation of an individual patch. However learning these representations independently is not optimal as pulmonary nodules would normally span a varying number of axial slices resulting in spatial dependencies that can be leveraged to learn better features. In order to address this aspect, the output of each CNN is passed on to an LSTM component. Each such component has a memory cell, which is modulated by a gating scheme determining the amount of information entering/leaving the cell. This memory cell allows the network to learn when to forget previous hidden states and when to update hidden states given new information. In our context, the role of the memory cells is to selectively preserve the amount of information extracted from each image patch, within the entire stack, in order to capture short- and long-range interdependencies. LSTM neurons have undergone many variations since its inception \citep{Hochreiter97}. The neuron we use here is described by the following functions (see also Figure~\ref{fig:lstm}),
 
 \begin{align}
{\bf i}_{j} & = \sigma({\bf W}_{iv} {\bf v}_{c,j} + {\bf W}_{ih} {\bf h}_{c,j-1} + b_{i}) \\
{\bf f}_{j} & = \sigma({\bf W}_{fv} {\bf v}_{c,j} + {\bf W}_{fh} {\bf h}_{c,j-1} + b_{f}) \\
{\bf o}_{j} &= \sigma({\bf W}_{ov} {\bf v}_{c,j} + {\bf W}_{oh} {\bf h}_{c,j-1} + b_{o}) \\
{\bf g}_{j} &= \phi({\bf W}_{gv} {\bf v}_{c,j} + {\bf W}_{gh} {\bf h}_{c,j-1} + b_{g}) \\
{\bf c}_{c,j} &= {\bf f}_{j} \odot {\bf c}_{c,j-1} + {\bf i}_{j} \odot {\bf g}_{j} \\
{\bf h}_{c,j} &= {\bf o}_{j} \odot \phi({\bf c}_{c,j}),
\end{align}
where $\sigma$ and $\phi$ are the sigmoid and hyperbolic tangent non-linearities, respectively, and denotes element-wise multiplication.

All gates ${\bf i}_{j}$, ${\bf f}_{j}$, ${\bf o}_{j}$, ${\bf g}_{j} \in \mathcal{R}^{B}$ of the $j$-th LSTM neuron receive as input the CNN representation ${\bf v}_{c,j}$ of the image patch ${\bf X}_{c,j}$ and the hidden state of the previous LSTM neuron in the sequence ${\bf h}_{c,j-1} \in \mathcal{R}^{B}$. Each gate, indexed by $k$, has its own weight matrices, i.e. ${\bf W}_{kv} \in \mathcal{R}^{B \times Q}$ and ${\bf W}_{kh} \in \mathcal{R}^{B \times Q}$, and a corresponding bias term $b_{k}$. The input gate ${\bf i}_{j}$ regulates the degree to which the input representation ${\bf v}_{c,j}$ would enter the memory cell to leverage its internal state ${\bf c}_{c,j} \in \mathcal{R}^{B}$. The forget gate ${\bf f}_{j}$ controls the contribution of the cell to the current hidden state by regulating the previous cell state ${\bf c}_{c,j-1} \in \mathcal{R}^{B}$. This gate can selectively prevent the current memory cell ${\bf c}_{c,j}$ to further propagate information from previous image patches representations ${\bf v}_{c,j-l}$ stored in previous memory cells ${\bf c}_{c,j-l}$. The input modulation gate ${\bf g}_{j}$ represents the input information that could enter the memory cell and it is modulated by the input gate. In our context, it represents the new information from the image patch representation ${\bf v}_{c,j}$, that could be stored into the memory cell ${\bf c}_{c,j}$. Finally, the output gate ${\bf o}_{j} \in \mathcal{R}^{B}$ controls the effect of the memory cell on the other LSTM neurons in the sequence. The property of the LSTM gating scheme to independently read, store and delete information from the memory cell allows ReCTnet to attend specific patches within the stack while downgrading the importance of others. This results in a mechanism able to learn complex interdependencies across adjacent image patches with the potential to improve the overall classification performance, as demonstrated in Section \ref{results}.

Following the principle that a deeper model can be more efficient at representing some functions than a shallow one (e.g. \cite{Bengio09}), we add two LSTM layers by stacking them on top of each other. We expect the higher LSTM layer to help capture more complex dependencies in the input sequence. Both the CNN and LSTM components use the same parameterization ${\bf K}$ and ${\bf W}$ across the mapping of each representation in the sequence, forcing the model to learn generic dependencies between consecutive image patches. This strategy prevents the parameter size from growing in proportion to the maximum number of adjacent image patches.  The output ${\bf h}_{c,j}$ of each neuron of the last LSTM layer is concatenated into a single vector ${\bf z} \in \mathcal{R^{P}}$ to obtain the representation of the whole input sequence. This representation is passed as input to a multi-layer perceptron (MLP) consisting of $2$ hidden layers; this produces a higher-order representation that is more easily separable into the two different classes. The activation of the $m$-th unit of the first hidden layer is given by
\begin{equation}
u^{(1)}_{m} =  f(\sum_{i = 1}^{P} q^{(1)}_{i,m}z_{i} + b^{(1)}_{j}),  
\end{equation}
where $f$ represents a non-linear activation function, $z_{i}$ is the $i$-th unit of the vector ${\bf z}$, $q^{(1)}_{i,m}$ is the connection weight between the unit $z_{i}$ and the unit $u^{(1)}_{m}$ of the first hidden layer. The connection weights of both MLP hidden layers are denoted by ${\bf Q}$. In its general form, the MLP is parameterized by ${\bf Q}$ and maps the concatenated sequence of the LSTM neurons outputs into a fixed length vector ${\bf u}_{c,j} \in \mathcal{R}^{D}$. Finally, the output of the MLP is passed as input to a softmax function to estimate the probability that the sampled voxel $c$ belongs to a nodule,
\begin{equation}
p(\hat{y}_{c,j} = \text{\enquote{nodule}}| {\bf u}_{c,j}; {\bf r}_{1}, {\bf r}_{2}) = \frac{\exp\{{\bf r}_{1}{\bf u}_{c,j}\}}{\sum_{k=1}^{2} \exp\{{\bf r}_{k}{\bf u}_{c,j}\}},
\end{equation}
where the vectors ${\bf r}_{1} \in \mathcal{R}^{D}$ and ${\bf r}_{2} \in \mathcal{R}^{D}$ are the columns of the softmax matrix ${\bf R}_{D \times 2}$. The resulting architecture is parameterized by the weights $({\bf K}, {\bf W}, {\bf Q}, {\bf R})$, which are learned jointly by minimizing the negative log-likelihood function,

\begin{equation}
\ell = - \sum_{c = 1}^{n} \log P_{{\bf K}, {\bf W}, {\bf Q}, {\bf R}} ( \hat{y}_{c,j} = y_{c,j} | \mathcal{X}_{c, j})
\end{equation}
through stochastic gradient descent with mini batches (MSGD) \citep{Bousquet08} and backpropagation.
\subsubsection{Automated object detection through probability maps}\label{validation}

\begin{figure}[t]
\begin{center}
\includegraphics[width=5.2in]{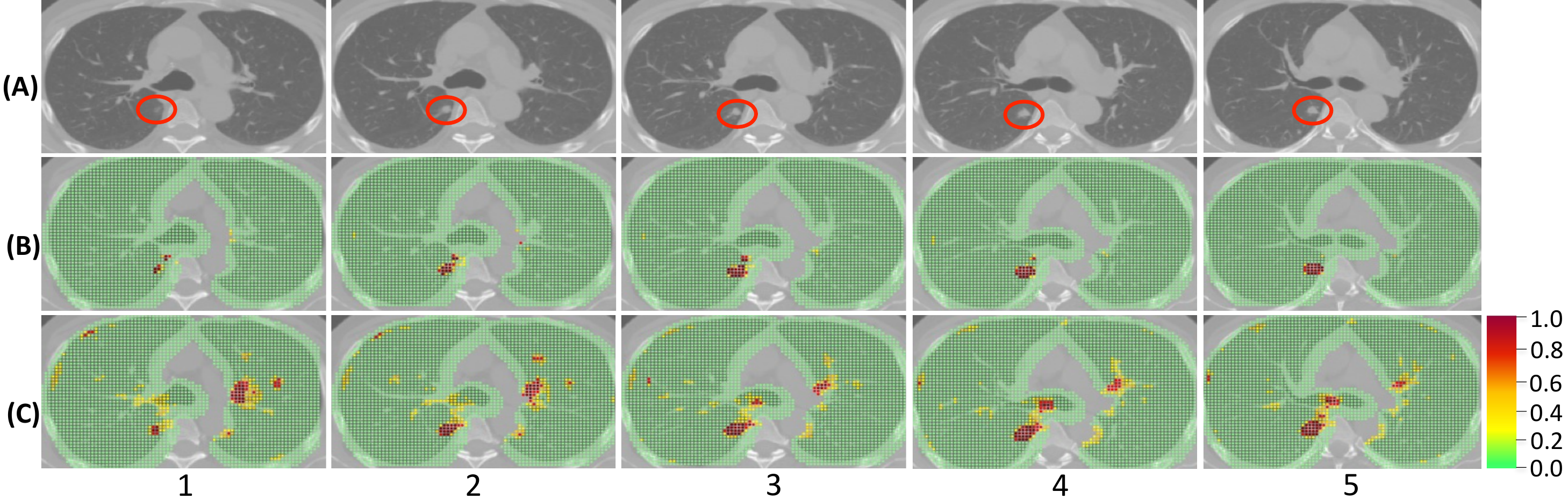}
\end{center}
\caption{(A) A sequence of consecutive axial slices containing the lung area with cross-sections of a nodule circled in red; (B) Corresponding probability maps generated by ReCTnet; (C) Corresponding probability maps generated by the CNN. It can be observed that ReCTnet assigns high probability to localized regions whilst the maps produced by CNN result in a larger number of false positives.}
\label{fig:maps}
\end{figure}

The detection of nodules in an unseen CT scan ${\bf S}$ is carried out in three steps. First, we segment the lungs using the segmentation algorithm described in Section \ref{lung_detection}. Rather than classifying all voxels contained in the lung area, we sample a subset of voxels whilst ensuring that we achieve enough coverage to detect small nodules. For all our experiments we have used  a uniform sampling strategy, as described in Section \ref{sampling1}, with sampling distance of $4$ times the in-plane pixel length of the corresponding scan in both grid directions. For each candidate voxel, a stack of rectangular patches is generated as described in Section \ref{sampling1}, which is then passed to ReCTnet for classification. This yields a probability map, $\mathcal{M}_{\bf{S}}$, indicating the likelihood that each sampled voxel belongs to a nodular area (e.g. see Figure~\ref{fig:maps}). 


\begin{figure}[t]
\begin{center}
\includegraphics[width=5.0in]{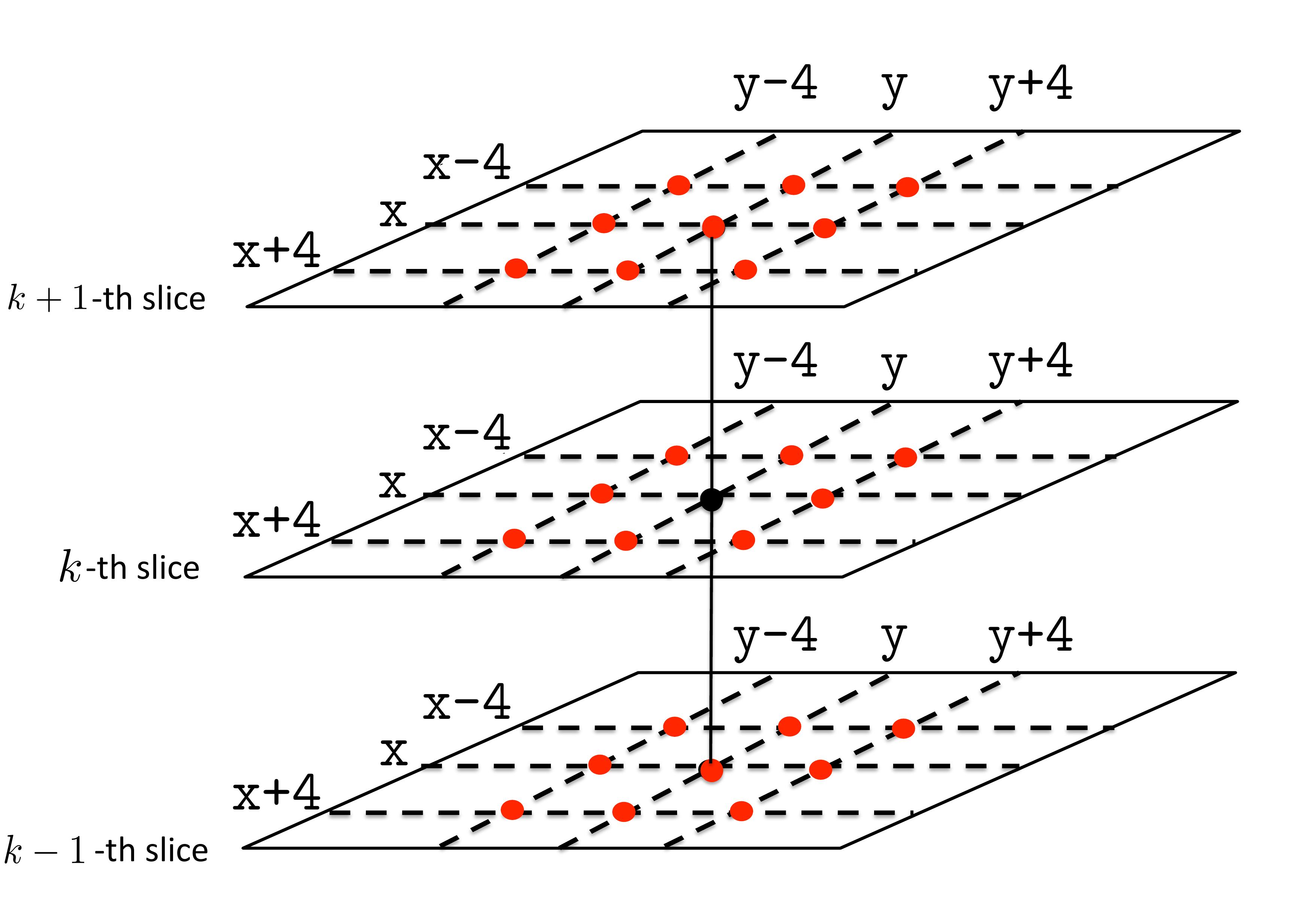}
\end{center}
\caption{Illustration of neighbourhood $N\{{\bf a}_{i}\}$ of a voxel with coordinates $(\texttt{x},\texttt{y})$ in the $k$-th slice (black point). The red points represent the neighbouring voxels of the voxel with coordinates $(\texttt{x},\texttt{y})$ within the same $k$-th slice and across the $k-1$-th and $k+1$-th adjacent slices.}
\label{fig:neighbourhood}
\end{figure}

\begin{figure}[t]
\begin{center}
\includegraphics[width=5.2in]{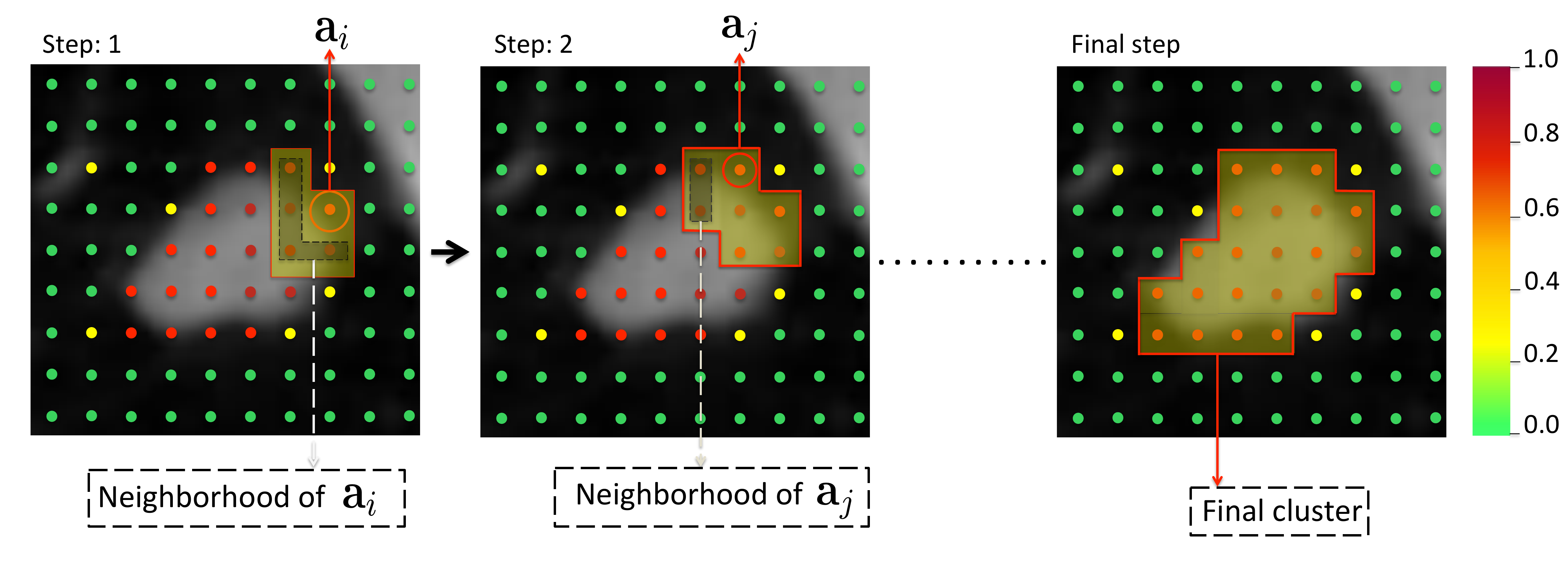}
\end{center}
\caption{A two-dimensional example of the region growing procedure for nodule candidates formation. In Step 1, the algorithm randomly selects a voxel ${\bf a}_{i}$ among those with estimated nodule probability $p_{i} \ge 0.5$ and its neighbourhood $\mathcal{N}\{{\bf a}_{i}\}$ is identified. In Step 2, ${\bf a}_{j}$ is randomly selected from $\mathcal{N}\{{\bf a}_{i}\}$ and all its neighbours are added to the region. By iterating this procedure the region is expanded until no more voxels can be added.} 
\label{fig:cluster}
\end{figure}

In a final post-processing step, candidate nodules are generated from these probability maps. First, we filter out all the sampled voxels that have been assigned a nodule probability of less than $0.5$. All the remaining voxels are then collected into a tensor $\mathcal{A}_{{\bf S}} := [{\bf a}_{1}, \hdots , {\bf a}_{n}]$, where ${\bf a}_{i} \in R^{3}$ represents the $i^\text{th}$ voxel's coordinates. Our aim is to identify clusters of spatially contiguous voxels that will generate candidate nodules. In order to do this, we introduce the notion of a voxels's neighbourhood. For a given voxel ${\bf a}_{i}$, its neighbourhood $\mathcal{N}\{{\bf a}_{i}\}$ contains all the voxels directly adjacent to ${\bf a}_i$ in any direction (see Figure~\ref{fig:neighbourhood}). We apply an agglomerative algorithm that initially starts with a randomly selected voxel ${\bf a}_{i}$ and adds it to cluster $\mathcal{C}_{i}$. The cluster is then expanded by adding all the voxels in $\mathcal{N}\{{\bf a}_{i}\}$ as well as their own neighbours; the procedure continues iteratively until no more voxels can be added and the cluster cannot be further increased. Figure~\ref{fig:cluster} provides an illustration of this process in two dimensions. In order to detect other clusters, all the voxels in $\mathcal{C}_{i}$ are removed from $\mathcal{A}_{{\bf S}}$, and the cluster-forming procedure above is repeated iteratively, until no more clusters can be found.

Any cluster $\mathcal{C}_{i}$ detected by the above algorithm will contain a certain number of voxels with classification probabilities in the range between $0.5$ and $1$. We prefer to further remove groups of voxels that contribute to lowering the average probability within the cluster and increasing its variance. This is accomplished by further stratifying all voxels within a cluster according to their classification probabilities. Accordingly, we apply a mode-seeking and non-parametric procedure, the mean shift algorithm with a Gaussian kernel \citep{Fukunaga75}. For each mode identified by the algorithm, we compute the average classification probability. Within each cluster, all modes with an average probability succeeding a given threshold, $p>0.5$, are merged and treated as a potential candidate nodule while all other voxels are discarded. The $p$ parameter controls both sensitivity and FP rate.

\subsubsection{A competing CNN architecture}

\begin{figure}[t]
\begin{center}
\includegraphics[width=5.5in]{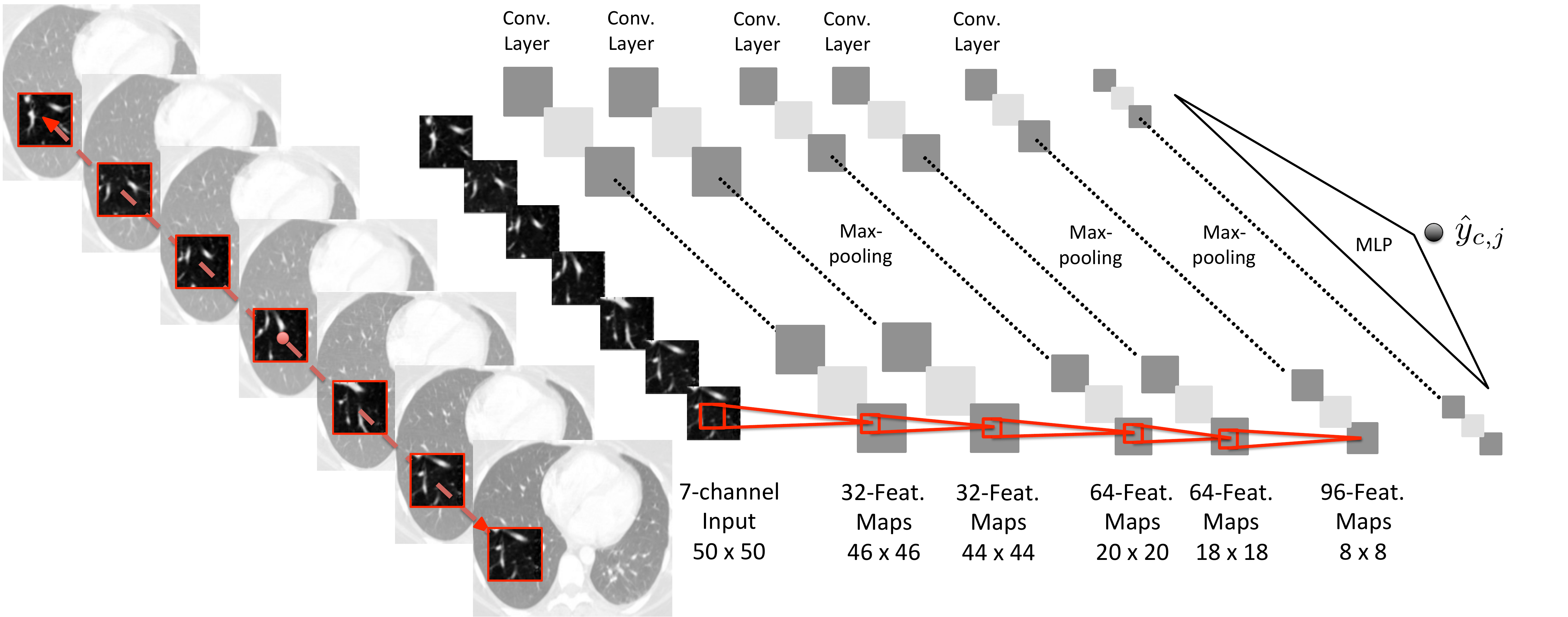}
\end{center}
\caption{An overview of CNNs architecture for learning dependencies within and across the adjacent CT image patches of the tensor $\mathcal{X}_{c,j}$ and use these dependencies to predict whether voxels belong to nodular or non-nodular areas. The CNNs architecture is composed from $5$ convolutional and $3$ max-pooling layers. The elements of the tensor $\mathcal{X}_{c,j}$ are used as input channels in order to form the feature maps in the first convolutional layer. All max-pooling layers are applied by $2 \times 2$ non-overlapped grids to down-sample each feature map by a factor of $2$ along each direction. The output of the last max-pooling layer is flattened and passed as input to MLP in order predict whether the voxel based on which the tensor $\mathcal{X}_{c,j}$ is formed belongs or not to a nodular area.}
\label{fig:cnn}
\end{figure}

The anatomical structures of healthy and nodular lung areas can be characterised by processing multiple adjacent CT slices. In this section we propose an alternative and simpler convolutional network without LSTM components that can also learn image representations by leveraging information across multiple slice. Similar fusing strategies have been utilized in different contexts, e.g. for video classification \citep{Karpathy14} and chemotherapy response prediction in multi-slice PET imaging \citep{Ypsilantis15}. This multi-channel CNN architecture (see Figure~\ref{fig:cnn}) will be used as benchmark in our experiments and provides the means to evaluate the benefits of adding recurrent layers as a mechanism to model inter-slice dependencies. 

For this model, the image patches in each tensor $\mathcal{X}_{c,j}$ are treated as input channels in the first convolutional layer. In this way, the CNN fuses the information across the entire sequence. A feature map ${\bf v}^{(1)}_m$ in the first convolutional layer is obtained by convolving each image patch ${\bf X}_{c,s}$ within the tensor $\mathcal{X}_{c,j}$ with a weight matrix ${\bf W}^{(1)}_{s,m}$, adding a bias $b_{m}^{(1)}$ and passing the result to a non-linear function $f$, 
\begin{equation}
{\bf v}_{m}^{(1)} = f(\sum_{s = j - k}^{j + k} {\bf W}^{(1)}_{s,m} \ast {\bf X}_{c,s} + b^{(1)}_{m}), \ \ \ m = 1, \hdots, 32.
\end{equation}
Each element of the $m$-th feature map ${\bf v}_{m}^{(1)}$ in the first convolutional layer encloses information from a local anatomical area of the lung as depicted across the adjacent image patches ${\bf X}_{c,s}$. The weight matrices ${\bf W}^{(1)}_{s,m} $, one for each feature map, are learned in order to build a library of low-level features that describe anatomical dependencies within each $2$D image patch as well as across multiple adjacent image patches.

\section{Experimental settings and results} \label{results}

\subsection{Implementation details}

In this section we provide the implementation details of both ReCTnet and the competing multi-channel CNNs. We introduce the following notation: we call I$(\phi)$ an input with $\phi$ channels, C$(\mu, \nu)$ convolutional layer with filters of size $\mu \times \mu$ and $\nu$ feature maps, P a pooling layer (with non-overlapping $2 \times 2$ pooling regions), and FC$(\kappa)$ a fully connected layer with $\kappa$ units. Using this notation, the CNN submodule of ReCTnet can be compactly described as: I$(1)$, C$(5,16)$, P, C$(4,16)$, C$(3,32)$, P, C$(3,64)$, P, FC$(412)$. 

ReCTnet is trained in two steps. First, we pre-train the architecture without the recurrent components by minimising the negative log-likelihood; at this stage we use single image patches of size $50 \times 50$ voxels, and by pass the output units of the FC$(412)$ layer on to a softmax function. Once this architecture has been fully trained, the softmax function is removed and the LSTM components are placed directly above the FC layer of the pre-trained CNNs. Each LSTM neuron has $612$ units per layer and its parameters are initialized uniformly in $[-0.1,0.1]$. The multi-layer perceptron (MLP) architecture has $2$ FC layers of size $1024$ and $512$ units, respectively. The competing multi-channel CNN is I$(7)$, C$(5,32)$, C$(3,32)$, P, C$(3,64)$, C$(3,64)$, P, C$(2,96)$, P, FC$(816)$, FC$(412)$. It has $5$ convolutional layers (C) and $3$ max-pooling layers (P). The output of the last max-pooling layer is flattened out and passed as input to MLP with $2$ hidden layers of size $816$ and $412$ units, respectively. The final number of parameters in ReCTnet and CNN is $10,691,950$ and $1,686,598$, respectively. Both architectures were trained on $600$ randomly selected scans and tested on $150$ scans from the LIDC/IDRI database. The remaining $268$ scans were used as validation set in order to tune the network parameters. 

All the activation functions are rectified linear units (ReLU) \citep{Nair10}, and SGD with momentum is used for training \citep{Sutskever13}. The initial learning rate is set to $0.05$ and it is halved during the training. The momentum coefficient is set to $0.7$ and remains stable during the training. Training took between $3$ and $5$ days using a single NVIDIA K40 GPU card. Processing a new CT scan takes on average $10$ minutes depending on its size, and the total number of sampled voxels ranges between $146$ to $800$ thousand. Our code is based on Torch $7$, a Lua library that supports GPU computing \citep{Collobert11}.

\subsection{Experimental results and comparisons to previous studies} 

Table~\ref{table:OtherCAD} summaries the performance results of the proposed neural network architectures alongside with analogous performance metrics that have been reported in previous published studies using the LIDC/IDRI dataset. We report on sensitivity, false positives (FPs) per scan, and the number of training and test scans. As it is generally desirable to reach high sensitivity on nodules that were identified with a high level of agreement between readers, for our architectures we report on nodules confirmed by all $4$ radiologists. 

A short overview of the competing methodologies is in order. The system developed by \cite{Messay10} initially identifies and segments candidate nodules by combining intensity thresholding and morphological features such as area, volume, circularity and sphericity of the nodule candidates. In the second stage, each candidate region is represented by a number of $2$D and $3$D imaging features quantifying the geometry, intensity and gradient of intensity. Linear and quadratic classifiers were used to identify nodules which resulted in a sensitivity of $82.7\%$ with $3$ FPs/scan on all nodules (agreement level $1$). In \cite{Tan11} a mean curvature feature combined with nodule and vessel enhancement filters were deployed to initially identify nodule candidates. In the second stage, a neural network classifier was trained on invariant morphological features describing the geometry of the nodule candidates and reached a sensitivity of $87.5\%$ with an average of $4$ FPs/scan on all nodules (agreement level $4$). \cite{Golosio08} used a multi-threshold surface triangulation to first identify nodule candidates. In the second stage, several features such as volume, roundness, maximum density, mass and principal moments of inertia were extracted from nodule candidates and provided as inputs to a fixed-topology artificial neural network. They obtained a sensitivity of $79.0\%$ with $4$ FPs/scan on all nodules (agreement level $4$). In \cite{Torres15}, a CAM algorithm \citep{Cerello10} was deployed to segment nodule candidates; shape and intensity features were then extracted and used as inputs to a neural network classifier achieving an $80.0\%$ sensitivity with $8$ FPs/scan on all nodules (agreement level $4$). In \cite{Teramoto12} a cylindrical filter was used to initially identify and enhance nodule candidates. In the second stage, features quantifying the shape of the enhanced nodule candidates were calculated and used as inputs to an SVM classifier achieving $87.0\%$ sensitivity with $4.2$ FPs/scan on nodules with diameter between $2$mm and $20$mm (agreement level $4$). 

\begin{figure}[t]
\begin{center}
\includegraphics[width=4.5in]{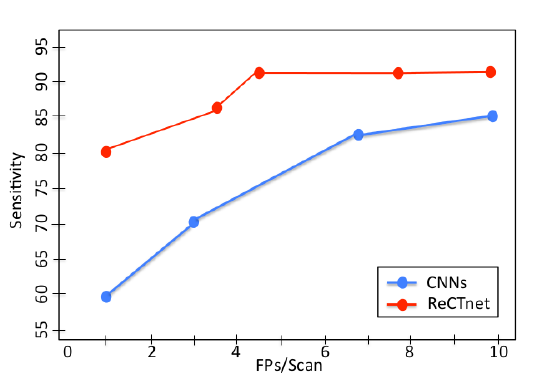}
\end{center}
\caption{FROC curves showing the performance of ReCTnet and CNNs on $150$ scans from LIDC/IDRI. As expected the nodule detection performance of both ReCTnet and CNNs decreases for lower values of FPs/Scan. Remarkably, ReCTnet achieves $80\%$ sensitivity with only one FPs/Scan, approximately $20.0 \%$ more of the sensitivity achieved by CNN.}
\label{fig:sensitivity}
\end{figure}

The results in Table~\ref{table:OtherCAD} suggest that the performance of ReCTnet compares well with other methods in terms of sensitivity at the given FP rates. In particular, ReCTnet reaches a detection rate of $90.5\%$ with $4.5$ FPs/scan (corresponding to a threshold of $p=0.75$) while lowering the FP rate to $3.5$ per scan yields a $85.6\%$ sensitivity. Compared to the simpler CNN architecture,  ReCTnet achieves higher identification rate for lower rate of FPs/scan; see Figure~\ref{fig:sensitivity}. It should also be noted that, with the exception of \cite{Torres15} where $949$ CT scans were used to validate the CAD system, we have used the largest validation set ($150$ test scans overall) compared to previous studies. Clearly, exact comparisons of performance metrics across these studies cannot be easily made for a number of reasons. First, the size and nature of both training and test dataset is often different. For instance, only $84$ scans were available in the LIDC database at the time when the first studies were published \citep{Armato04}. In some cases, as in \cite{Torres15}, the object detection algorithm was trained using $94$ cases, of which only $69$ where from LIDC/IDRI while the remaining $25$ were taken from two other sources; the performance was then tested on $949$ LIDC/IDRI scans. As noted, the detection targets (i.e. the agreement levels) for training and/or testing also vary across studies. The final performance results clearly depend on how these aspects are chosen. 

\begin{table}[t]
\centering
\caption{Performance comparisons of pulmonary nodule detection methods. $^{*}$ the algorithm was trained using additional datasets besides LIDC/IDRI. $\dagger$ the algorithm was tested using cross-validation.}\label{table:OtherCAD}
\scalebox{0.85}{
\begin{tabular}{c c c c c} 
\hline\hline
Study              & Sensitivity & FPs/scan &Training scans  & Test scans \\[0.5ex] %
\hline 
\cite{Golosio08}     & $79.0\%$ & $4.0$  & $84^{\dagger}$ & $84$ \\
\cite{Messay10}     & $82.7\%$ & $3.0$ &$84^{\dagger}$ & $84$ \\
 \cite{Tan11}         & $87.5\%$ & $4.0$ & $235$ & $125$ \\
\cite{Torres15}      & $80.0\%$  & $8.0$  & $94^{*}$  & $949$\\
\cite{Teramoto12}    & $87.0\%$  & $4.2$ & $84^{\dagger}$ & $84$ \\
\hline
ReCTnet & ${\bf 90.5} \%$ & ${\bf 4.5}$ & $600$ & $150$ \\
                             & $85.6 \%$ & $3.5$ & $600$ & $150$ \\
CNNs                      & $81.8\%$  & $6.8$ &  $600$ & $150$\\
                             & $70.0 \%$ & $3.0$ &  $600$ & $150$\\
\hline\hline   
\end{tabular}
}
\end{table}

\begin{figure}[t]
\begin{center}
\includegraphics[width=5.5in]{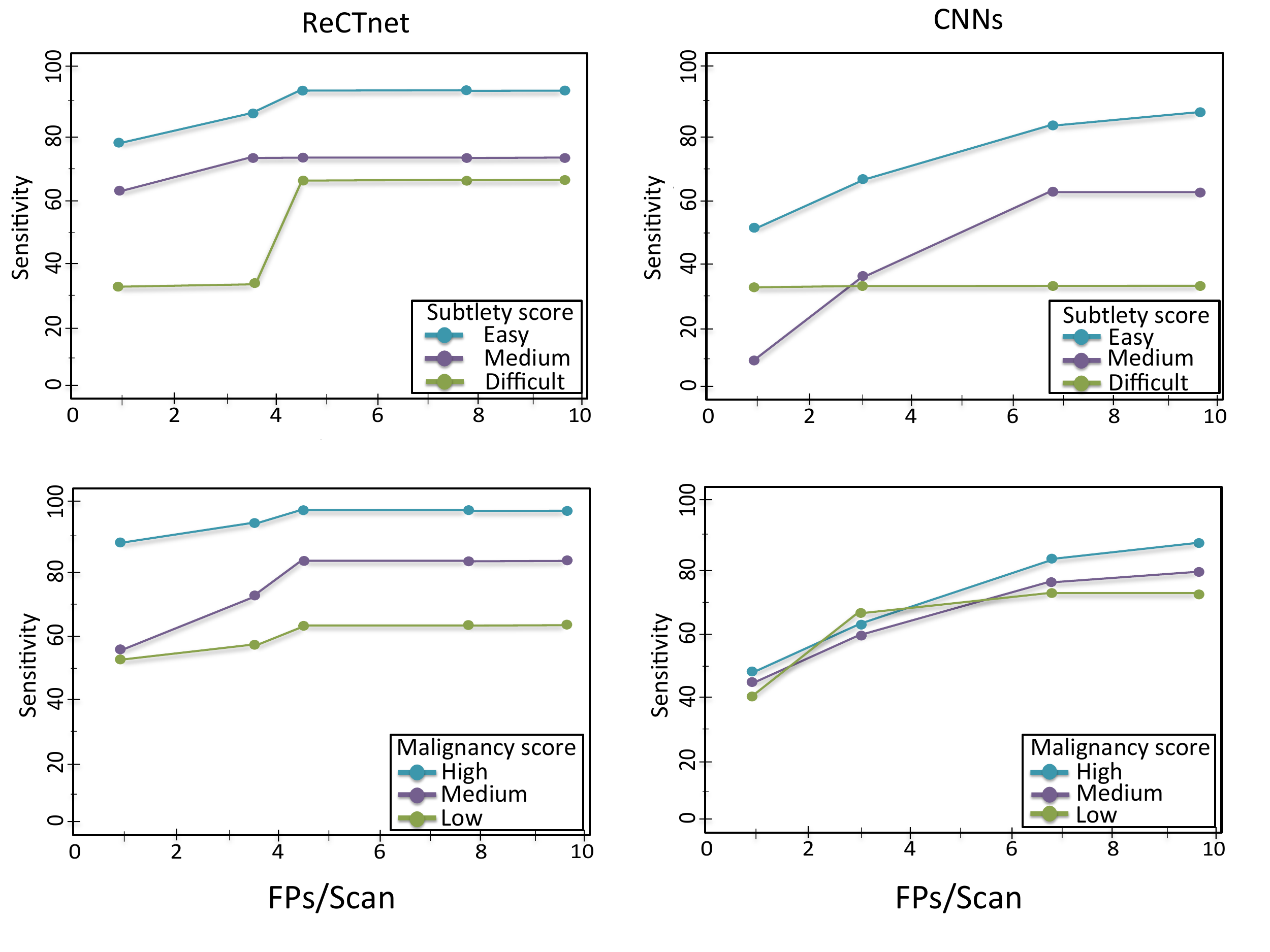}
\end{center}
\caption{FROC curves for ReCTnet (1st column) and CNNs (2nd column) by malignancy and subtlety scores. In general, ReCTnet outperforms CNNs. Also, the detection rate of CNNs decreases faster compared to the detection rate of ReCTnet for lower values of FPs/Scan.}
\label{fig:CNN_scores}
\end{figure}

In Table \ref{table:score_rats} we report on the detection sensitivity of both ReCTnet and CNN by subtlety and malignancy classes. As expected, the sensitivity rate achieved by ReCTnet increases proportionally to both subtlety and malignancy scores. Easy-subtlety nodules are detected with higher sensitivity ($92.0\%$ with $4.5$ FPs/scan) compared to medium-subtlety class nodules which are detected with sensitivity $73.0\%$, and difficult-subtlety class nodules detected with a lower sensitivity of $66.7\%$. For high malignancy nodules, ReCTnet achieves $96.5\%$ sensitivity with $4.5$ FPs/scan. Nodules with medium malignancy are identified with sensitivity $83.4\%$, and nodules with low malignancy are detected at $63.2\%$ sensitivity with $4.5$ FPs/scan. Figure~\ref{fig:CNN_scores} illustrates the FROC curves of both ReCTnet and CNN by subtlety and malignancy classes.  

Figure~\ref{fig:maps} provides some examples of probability maps generated by both ReCTnet and CNNs for five adjacent CT slices. As can be observed here, both models are able to place high probability to voxels falling in regions containing the nodules, and these high-probability voxels typically form spatially homogeneous clusters. However, CNN yields a substantially higher FP rate compared to ReCTnet. 

\begin{figure}[t]
\begin{center}
\includegraphics[width=5.0in]{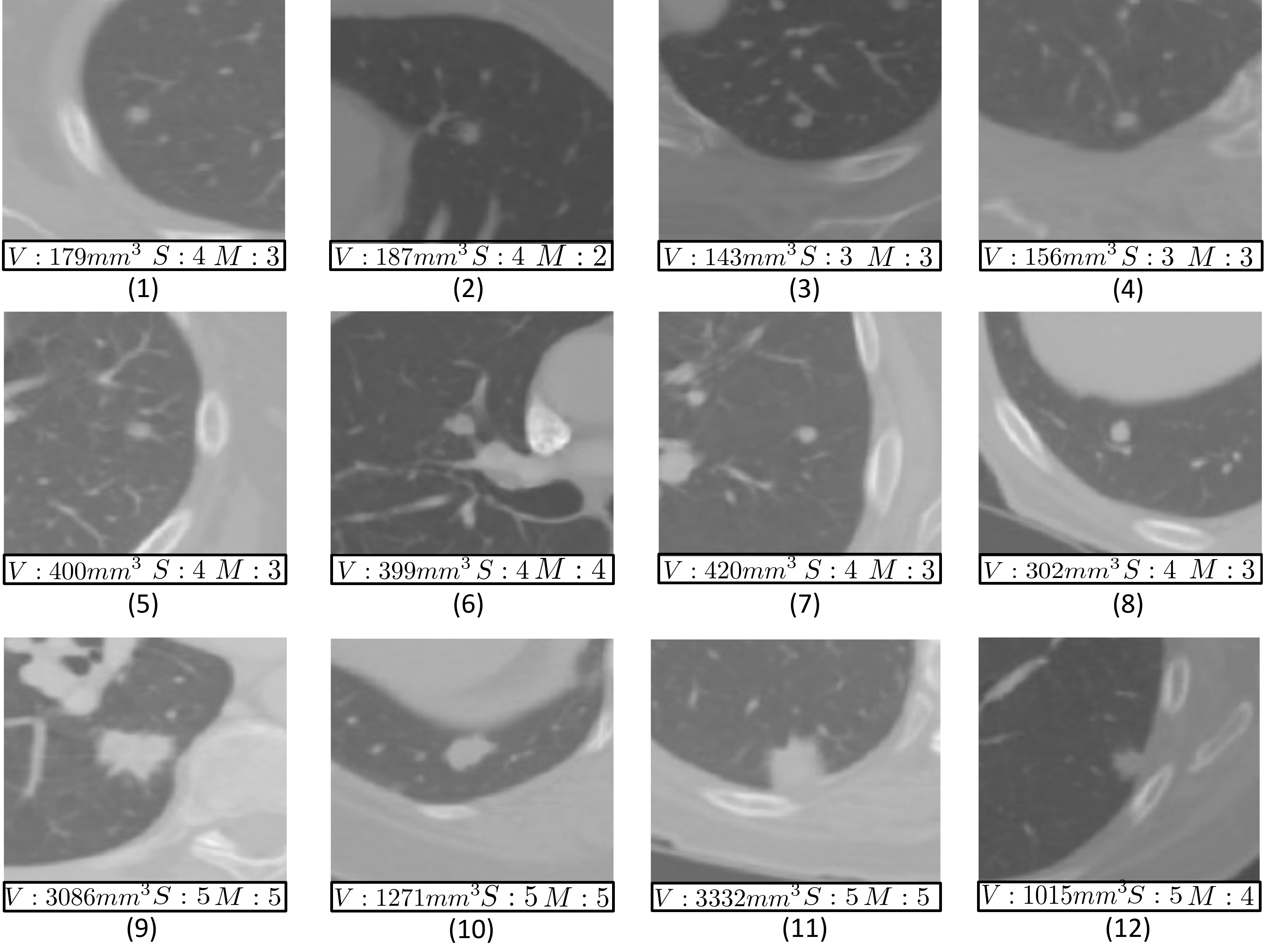}
\end{center}
\caption{Examples of lung nodules correctly detected by ReCTnet. All nodules are located in the centre of the image. We report on their volume $(V)$ in $mm^{3}$, subtlety score $(S)$ and malignancy score $(M)$.}
\label{fig:nodule_examples}
\end{figure}

\begin{figure}[t]
\begin{center}
\includegraphics[width=5.0in]{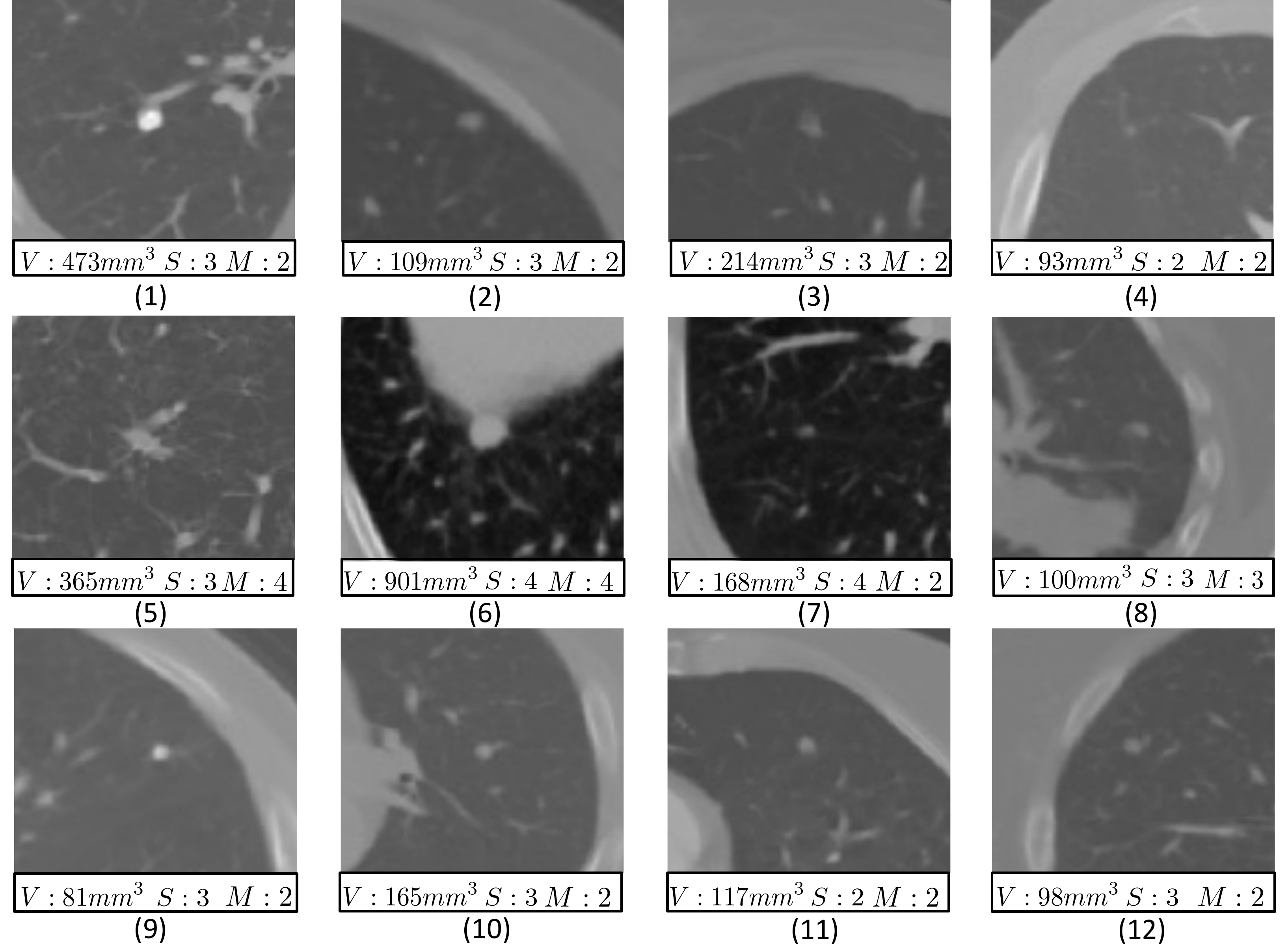}
\end{center}
\caption{Examples of pulmonary nodules that were not detected by ReCTnet. All nodules are located in the centre of the image. We report on their volume $(V)$ in $mm^{3}$, subtlety score $(S)$ and malignancy score $(M)$.}
\label{fig:undetected_examples}
\end{figure}

\begin{figure}[t]
\begin{center}
\includegraphics[width=5.0in]{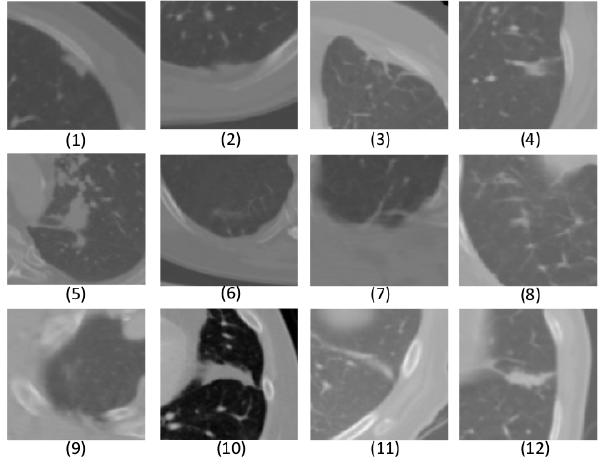}
\end{center}
\caption{Lung abnormalities detected by ReCTnet with $4.5$ FPs per scan and a sensitivity of $90.5\%$. The abnormalities are located in the centre of the images.}
\label{fig:detect_abnormalities}
\end{figure}

\begin{figure}[t]
\begin{center}
\includegraphics[width=5.0in]{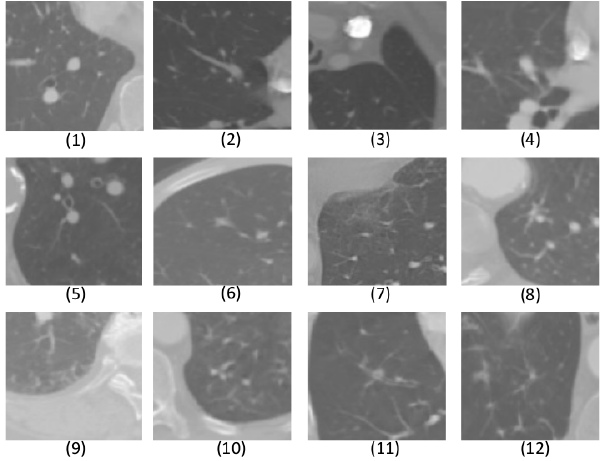}
\end{center}
\caption{FPs by ReCTnet with $4.5$ FPs per scan and a sensitivity of $90.5\%$. The FP are located in the centre of the images.}
\label{fig:false_positives}
\end{figure}

Figure~\ref{fig:nodule_examples} shows examples of individual slices containing pulmonary nodules that ReCTnet has been able to successfully detect. Examples of nodules that have been missed by ReCTnet are given in Figure~\ref{fig:undetected_examples} ; some of these cases appear to have low contrast while others are particularly difficult to detect due to their small size and the fact that the nodules are attached to normal pulmonary structures. Selected example of lung abnormalities detected by ReCTnet and false positives are illustrated in Figure~\ref{fig:detect_abnormalities} and Figure~\ref{fig:false_positives} respectively.

\begin{table}[t]
\centering
\caption{Performance comparisons by subtlety and malignancy ratings. The prediction sensitivity of ReCTnet and CNNs is reported for $4.5$ and $6.8$ FPs/scan respectively.} \label{table:score_rats}
\begin{tabular}{|c|c|c|c|c|}
\cline{1-4}
\multicolumn{2}{|c|}{\multirow{2}[4]{*}{ }} & \multicolumn{2}{c|}{Sensitivity} \\
\cline{3-4} 
\multicolumn{2}{|c|}{\multirow{2}[4]{*}{ }} & ReCTnet & CNNs \\
\cline{1-4}

{\multirow{3}{*}{Subtlety}} & Low  & $66.7\%$ & $34.0\%$ \\
                            & Medium &   $73.0\%$ & $64.0\%$ \\
                            & High  & $92.0 \%$ & $82.0\%$ \\
\hline

{\multirow{3}{*}{Malignancy}}  & Low &  $63.2\%$ & $73.3\%$ \\
                            & Medium  &   $83.4\%$ & $75.6\%$ \\
                            & High  & $96.5\%$ & $81.6\%$ \\
\cline{1-4}
\end{tabular}
\end{table}

\section{Discussion}\label{discusion}

First attempts to use CNNs as components of larger CAD systems for nodule detection in chest radiographs date back to the mid-nineties \citep{Lo95, Lin96}. However, early networks were shallow. More recently, deep residual networks, designed to ease the flow of the gradients during backpropagation \citep{He15}, have been proposed to identify nodules directly from raw chest radiographs \citep{Bush16}. For pulmonary nodule detection using CT imaging, CNNs have recently been used as a feature extractor within a larger CAD system \citep{Ginneken15}. In that work, upon identifying nodule candidates using more traditional image analysis techniques, selected image patches were extracted in the sagittal, coronal and axial plane centred at every possible location of the candidate regions and used as inputs in pre-trained CNNs to extract imaging features; support vector machines were then used for the final classification task. In other medical applications, CNNs using three orthogonal patches (triplanar CNNs) have been utilized for the detection of lymph nodes and polyps in CT scans \citep{Roth14, Roth15}. LSTMs' ability to explore the context of each voxel has also recently been proposed for medical image segmentation using fully-volumetric images \citep{Stollenga15}. Outside of the medical imaging arena, the combination of CNNs and RNNs has been explored to represent video sequences and implicitly learn spatio-temporal features for both video recognition and description \citep{Donahue14, Venugopalan14}.

During the development of ReCTnet, a number of alternative architectural modifications and training methods were tested, but did not result in substantial improvements. We implemented a deterministic attention mechanism, somewhat similar to the approach described in \cite{Bahdanau14}, in order to learn a weighted sum of LSTM outputs and use it as input to the MLP. Our rationale was to introduce an additional mechanism to automatically down-weight the importance of certain image patches within an input sequences. However, this modification did not yield any substantial improvements in performance, indicating that the gates of the LSTM neurons were able to attend to specific parts of the input sequence while ignoring others through independently reading, writing and erasing content from the memory cells. Furthermore, to determine whether pre-training the CNN components without any additional fine tuning was sufficient, we trained only the LSTMs and MLP parameters using as inputs the fixed-length representation of the pre-trained CNN layers. This approach gave lower performance compared to fine tuning, thus demonstrating the importance of end-to-end training. At the same time, we also assessed the impact of having more than one LSTM layers. The 2-layer architecture reported here resulted in higher performance, compared to a single layer, demonstrating that adding depth in this context introduces additional flexibility in capturing complex spatial dependencies in the input sequences. 

The effect of regularisation was tested through a dropout mechanism applied to non-recurrent connections of the LSTM architecture \citep{Zaremba14}. In previous studies, dropout has been found to reduce overfitting by preventing the LSTM units from co-adaptations \citep{Srivastava14}. However, in our case this strategy did not improve the performance in any substantial way. 

Finally, we experimented with an adaptive sampling strategy for the automated selection of training voxels from raw scans directly at training time. Starting with a uniform sampling grid, during training we periodically (every $5$ epochs) increased the sampling density through the addition of a certain number of voxels associated with high prediction errors. The rationale was to boost the generalization ability of the model by training it on an dynamically increasing number of input data points that appeared to be particularly difficult to classify. However, in our experiments, the adaptive sampling strategy did not improve the performance compared to uniform sampling. 

\section{Conclusion}\label{conclusion}
%
%

In this study we have presented a deep neural architecture that learns to identify anatomical objects of interests in CT scans. ReCTnet has been tested on the challenging task of nodule identification in thoracic CT scans using a large publicly available dataset and has achieved detection sensitivity  that is generally higher than previously proposed methodologies. We have found that our architecture also outperforms a competing multi-channel CNN thus indicating the important role played by the recurrent components used by ReCTnet. Our results suggest that the LSTM layers enable to better synthesize the anatomical information across adjacent slices eventually resulting in improved discrimination ability. 

Besides the application presented here, the proposed architecture is sufficiently flexible to be used for other object detection tasks involving different imaging modalities, e.g. magnetic resonance imaging and positron emission tomography, characterised by inter-slice dependencies. The network can be trained end-to-end from raw image patches. Its main requirement is the availability of a sufficiently larger training database, but otherwise no assumptions are made about the objects of interest or underlying image modality. Clearly, some of the architectural choices that have been made for this application may not always be optimal and would need to be reconsidered. In particular, the use of $7$ adjacent slices to represent the anatomical context of a voxel might not always yield best performance. In future work, we would like to develop an extension to automatically identify how much context is required for each training voxel, i.e. how many adjacent slices should be included during training. We are also planning to develop an attention mechanism, using reinforcement learning, to automatically identify which portions of a scan would contain the relevant information thus removing the need to perform the initial lung segmentation process.

\vspace{10pt}

\bibliography{ct_refs}

\end{document}